# Segmentation-Aware Generative Reinforcement Network (GRN) for Tissue Layer Segmentation in 3-D Ultrasound Images for Chronic Low-back pain (cLBP) assessment

Zixue Zeng [1, 2], Xiaoyan Zhao [1], Matthew Cartier [3], Tong Yu [1], Jing Wang [1], Xin Meng [1], Zhiyu Sheng [4], Maryam Satarpour [1], John M Cormack [4], Allison Bean [5], Ryan Nussbaum [5], Maya Mauer [6], Emily Landis-Walkenhorst [6], Dinesh Kumbhare [7], Kang Kim [1, 4], Ajay Wasan [6, 8], Jiantao Pu [1, 2, 9,*]

[1] Department of Bioengineering, University of Pittsburgh, Pittsburgh, PA 15213, USA

[2] Department of Radiology, University of Pittsburgh, Pittsburgh, PA 15213, USA

[3] Department of Mathematics, University of Pittsburgh, Pittsburgh, PA 15213, USA

[4] Cardiology, Department of Medicine, University of Pittsburgh

[5] Department of Physical Medicine and Rehabilitation, Pittsburgh, PA 15213, USA

[6] Anesthesiology & Perioperative Medicine, School of Medicine, University of Pittsburgh

[7] Department of Medicine, Division of Physical Medicine and Rehabilitation, University of Toronto, Toronto, Ontario, Canada.

[8] Psychiatry, School of Medicine, University of Pittsburgh

[9] Department of Ophthalmology, University of Pittsburgh, Pittsburgh, PA 15213, USA

**Acknowledgement:** This work is supported in part by research grants from the National Institutes of Health (NIH) (R61AT012282 and R01CA237277).

## Abstract

Layer-wise segmentation of three-dimensional (3D) ultrasound for chronic lower back pain (cLBP) requires large amount of labeled images. To mitigate this burden, we propose a Generative Reinforcement Network (GRN) that integrates a generative adversarial network (GAN) framework with a segmentation model: a generator is coupled to a segmentor via segmentation-aware feedback while also being regularized by a discriminator. At each iteration, the segmentation loss is back-propagated into the generator so it learns to produce denoised, easy-to-learn reconstructions that directly reduce downstream segmentation error (reinforcement augmentation, RAug), and adversarial feedback from the discriminator (PatchGAN) further encourages realistic reconstructions. We further introduce segmentation-guided enhancement (SGE), in which the pre-trained generator enhances input images at the inference stage to improve segmentation. We proposes two variants: GRN-SEL, which uses RAug only, and GRN-SSL, which additionally applies interpolation-consistency training (ICT) on unlabeled data by interpolating generator-reconstructed pairs and enforcing prediction consistency on the interpolated images. We evaluate GRN on a fully annotated lumbar back ultrasound dataset (LBUD) and two external datasets. On the LBUD dataset, GRN-SEL with SGE reduces labeling efforts by up to 70% while achieving a 1.98% improvement in the Dice Similarity Coefficient (DSC) compared to models trained on fully labeled datasets. Additionally, across all three datasets and label fractions, GRN variants consistently outperform state-of-the-art semi-supervised methods. These findings suggest the effectiveness of the GRN framework in optimizing segmentation performance with significantly less labeled data, offering a scalable and efficient solution for ultrasound image analysis and reducing the burdens associated with data annotation. The code is publicly available at https://github.com/Francisdadada/GRN.

**Key Words**

# 1. Introduction

Chronic lower back pain (cLBP) is a complex and multifaceted condition [1]. It is highly prevalent, with 70-85% of individuals experiencing lower back pain (LBP) at some point in their lives [2]. Additionally, 19.6% of people aged between 20 and 59 report LBP [3, 4], significantly affecting their quality of life [5]. Despite extensive research, etiology-based diagnosis and successful management of cLBP remain elusive, with no consensus on the most effective approaches [6].

Various clinical imaging techniques, including ultrasound, computed tomography (CT), and magnetic resonance imaging (MRI), have been employed to explore the underlying classification and mechanisms of cLBP. CT provides high-resolution cross-sectional images, offering detailed visualization of various anatomical structures, including the vertebrae and muscles [7-9]. However, CT imaging exposes patients to ionizing radiation, which carries potential long-term health risks. Also, it has limited ability to differentiate between soft tissues, making it less effective for detailed assessment of muscular structures [10]. MRI is a non-invasive diagnostic tool that uses powerful magnetic fields and radio waves to produce high-resolution images, particularly suited for visualizing soft tissues, organs, and nervous systems [11, 12]. However, MRI imaging is generally more expensive, less accessible, and time-consuming compared to other imaging methods, limiting its use. By comparison, ultrasound imaging offers a cost- and time-effective alternative without the risks associated with ionizing radiation, making it a valuable and widely used tool in clinical practice for assessing cLBP.

Current ultrasound research on cLBP employs predominantly two-dimensional (2-D) ultrasound imaging to assess muscle thickness, contraction, and tissue echogenicity, which are believed to be associated with LBP[13]. This 2-D imaging approach overlooks volumetric and intricate anatomical structures and thus limits the comprehensiveness of the analysis [14]. In contrast, three-dimensional (3-D) ultrasound imaging offers improved spatial visualization, enabling a more detailed and accurate layer-by-layer analysis from the dermis to the muscle. This comprehensive volumetric data allows for the identification of subtle morphological changes and the assessment of tissue interactions, making 3-D ultrasound a superior modality for studying cLBP. However, each 3-D scan can produce hundreds of image slices, making manual annotation extremely labor-intensive and limiting the availability of annotated data necessary for training robust deep learning models. Also, this technique may be constrained by limited penetration depth and is prone to artifacts and noise, which can compromise image quality[15]. Furthermore, existing studies typically focus on a limited number of tissues rather than performing a comprehensive layer-by-layer analysis from the dermis to the muscle. Consequently, essential information regarding individual tissues and their interactions across layers might be omitted. Therefore, an automated layer-by-layer segmentation solution would be highly valuable for identifying image biomarkers associated with cLBP from 3-D ultrasound imaging data.

Several advanced machine learning techniques are being explored to address the challenge of training AI modes with limited annotated data, such as few-shot learning[16], sample-efficient learning (SEL), Semi-supervised learning (SSL)[17], and Generative Adversarial Networks (GANs) [18]. Few-shot learning emphasizes rapid generalization to new tasks rather than improving performance on existing tasks with limited data. In contrast, SEL aims to maximize model performance while minimizing training examples. A common SEL strategy is data augmentation [19]. Some investigators have advocated for augmentation methods that generate challenging examples by maintaining an adversarial relationship with downstream models to minimize task-specific losses to enhance robustness [20]. Others focus on minimizing task-specific losses by optimizing augmentations to produce easy-to-learn samples and thus improve performance [21]. However, limited research has explored the potential of developing a more comprehensive augmentation framework that generates both challenging and easy-to-learn samples. SSL combines supervised training on a limited amount of labeled data with unsupervised training on a larger pool of unlabeled data to enhance performance and generalization [17, 22-26]. However, SSL depends on high-quality labeled data and is based on key assumptions, such as smoothness (similar data points yield similar predictions) and clustering (data naturally forms distinct and well-defined groups), which may not always be valid in real-world scenarios. GAN has been widely used to generate synthetic images to augment data [27-30]. However, GAN-based augmentation typically requires pre-training the model to generate synthetic images. These images may not accurately replicate real-world images and do not always lead to improved performance for segmentation models [31].

In this study, we propose an innovative framework termed generative reinforcement network (GRN). GRN integrates GAN model and segmentation models (segmentor) through a collaborative learning loop. Unlike traditional data augmentation techniques, GRN establishes a feedback mechanism where the segmentation model actively guides the generator during each training iteration via back-propagated losses. This design ensures that the generator produces easy-to-learn augmented images specifically optimized to improve segmentation performance. The discriminator provides adversarial signals to the generator, ensuring that the augmented images appear realistic. Additionally, interpolation consistency training is employed to generate challenging samples to enhance the model's robustness when applied to unseen data. During inference, the generator also functions as an image enhancement tool for preprocessing inputs by emphasizing features critical for segmentation, thereby improving the reliability of tissue segmentation. GRN supports both SEL and SSL scenarios. By reducing the dependence on extensive manual labeling, it achieves competitive and often superior performance compared to fully supervised models. This efficiency makes GRN particularly effective for problems like ours, namely tissue layer segmentation in 3-D ultrasound imaging.

## 2. Materials and Methods

### 2.1 Motivation

Curriculum learning employs a scheduler that repeatedly picks the easiest samples for the segmentor, promoting a smoother and more stable training process. Inspired by this, GRN replace the scheduler with generator that could transform the challenging images into easy ones. To ensure the generator enhance the images as easy-to-learn images, we utilize a feedback mechanism where the segmentor loss is backpropagated to the generator during each training iteration. This approach, termed reinforcement augmentation (RAug), optimized the following objective function:

$$min_G \left( min_S L_{seg}(S(G(x)), y) \right), where\ x \sim p_{data}(x) \quad (1)$$

where $x$ is sampled from the prior distribution $p_{data}(x)$, and $y$ represents the ground truth. $G$ and $S$ denote the generator and segmentor, respectively. $L_{seg}$ is the segmentor's loss.

We propose two GRN variants: **GRN-SEL**, which focuses on SEL using only RAug, and **GRN-SSL**, which integrates interpolation consistency training for SSL.

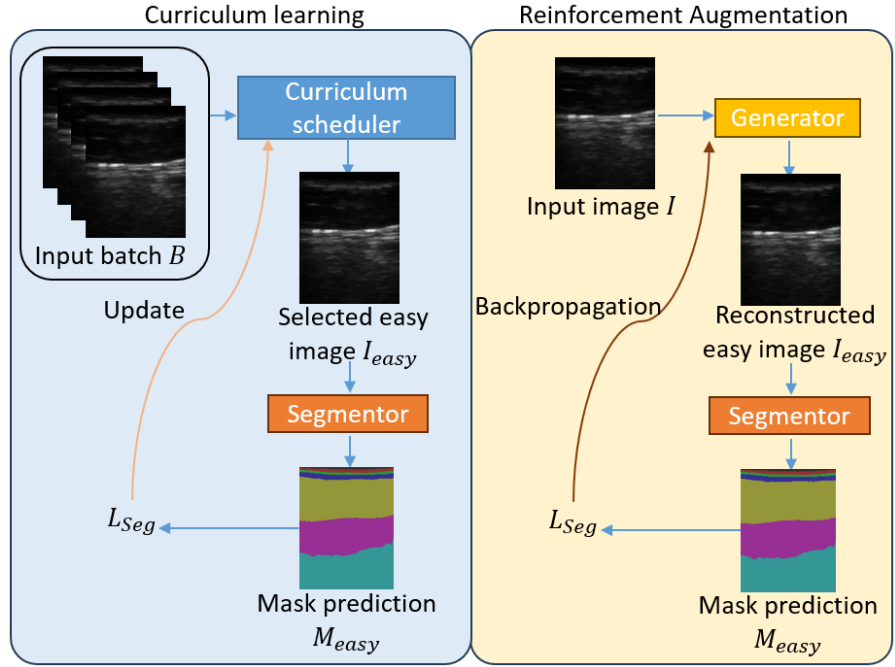

**Fig. 1.** Comparison between curriculum learning and GRN. In curriculum learning, a scheduler picks the easiest images for the segmentor, whereas GRN replace this scheduler with a generator that transforms inputs into simpler images, accelerating model convergence.

### 2.2 Generative Reinforcement Network

### 2.2.1 Sample-efficient and Semi-supervised Generative Reinforcement Network

GRN-SEL combines a GAN model with the segmentor feedback mechanism (Fig. 2). In each training iteration, the generator $G_\theta$ processes each labeled image to produce a reconstructed version. The segmentor $S_{\theta''}$ then generated predicted masks from those reconstructed images. A Dice loss is calculated based on the difference between these predicted masks and the ground truth masks, and the loss is back-propagated to the generator. This feedback encourages $G_\theta$ to refine its reconstruction by removing noise and irrelevant details, enabling the segmentor to focus on important features and thereby improving robustness.

Like a traditional GAN model, the generator $G_\theta$ also receives feedback from the discriminator $D_{\theta'}$, which classifies each reconstructed image as real or fake. Additionally, the segmentor processes both original, unaltered images and the reconstructed images and calculates a combined Dice loss based on mask prediction from both. This dual-processing approach promotes the segmentation model's robustness and stability by ensuring consistent performance on both generator-processed and original images. The comprehensive loss functions are defined as follows:

$$L_G = \lambda_{adv} L_{MSE}\big(D_{\theta'}(\widehat{I_L}), 1\big) + \lambda_{seg} L_{Dice}\big(\widehat{M_L}, M\big) \\ + \lambda_{L1} L1\big(\widehat{I_L}, I_L\big) \tag{2}$$

$$L_S = \frac{L_{Dice}(M_L, M) + L_{Dice}\big(\widehat{M_L}, M\big)}{2} \tag{3}$$

$$L_D = \frac{L_{MSE}\big(D_{\theta'}(I_L), 1\big) + L_{MSE}\big(D_{\theta'}(\widehat{I_L}), 0\big)}{2} \\ // \ 1: \text{Real}, 0: \text{Fake} \tag{4}$$

where $L_{MSE}$ represents the Mean Squared Error (MSE) loss. $L_{Dice}$ represents the Dice loss and $L1$ represents the L1 loss. $\lambda_{adv}$, weights $\lambda_{seg}$ and $\lambda_{L1}$ control the contributions to each loss component. $I_L$ and $\widehat{I_L}$ are the labeled image and its reconstructed version. $M$ is the ground truth mask. $M_L$ and $\widehat{M_L}$ are the mask predictions for the labeled image and reconstructed image, respectively. The generator loss function combines both generator and discriminator feedback loss terms, ensuring image generation supports not only realism but also effective learning by the segmentor.

GRN-SSL extends the segmentor feedback mechanism to the unlabeled training process by replacing the segmentation loss with a customized loss that evaluates segmentation model consistency (i.e., interpolation consistency loss in this study) (Fig. 2). A two-phase alternate-batch strategy is implemented to balance labeled and unlabeled datasets. In each iteration, a labeled $B_L$ and an unlabeled batch $B_{UL}$ are sampled from the labeled dataset $Dataset_L(I, M)$ and the unlabeled dataset $Dataset_{UL}(I, M)$, respectively. The training process includes two phases for each iteration, a supervised training phase with $B_L$ and an unsupervised training phase with $B_{UL}$. The supervised training phase mirrors the process in GRN-SEL, except that discriminator training is moved to the unsupervised training phase. Since $Dataset_{UL}(I, M)$ contains significantly more images than $Dataset_L(I, M)$, including the discriminator in the unsupervised training is sufficient to enhance model robustness. The loss functions are as follows:

$$L_G = \lambda_{seg} L_{Dice}(\widehat{M_L}, M) + \lambda_{L1} L1(\hat{I}_L, I_L) \tag{5}$$

$$L_S = \frac{L_{Dice}(M_L, M) + L_{Dice}(\widehat{M_L}, M)}{2} \tag{6}$$

In the unsupervised training phase, the generator processes unlabeled images to produce reconstructed versions, and the segmentor generates the corresponding masks. Instead of using traditional segmentation loss like GRN-SEL, a customized consistency loss $L_{customize}$ is defined and back-propagated to the generator. The customized consistency loss can be any predefined loss function used to evaluate model robustness. In this study, we specifically use interpolation consistency loss. As in GRN-SEL, the discriminator evaluates the realism of the generated images and provides feedback to guide image generation. The overall loss function for phase 2 is defined as follows:

$$L_G = \lambda_{adv}L_{MSE}\left(D_{\theta'}\left(\widehat{I_{UL}}\right), 1\right) + \lambda_{cus}L_{Customize}\left(\widehat{M_{UL}}, M_{UL}\right)$$
$$+\lambda_{L1}L1\left(\widehat{I_{UL}}, I_{UL}\right) \tag{7}$$

$$L_S = \frac{L_{customize}\left(\widehat{M_{UL}}\right) + L_{customize}(M_{UL})}{2} \tag{8}$$

$$L_D = \frac{L_{MSE}(D_{\theta'}(I_{UL}), 1) + L_{MSE}\left(D_{\theta'}\left(\widehat{I_{UL}}\right), 0\right)}{2}$$
$$// \; 1: \text{Real}, 0: \text{Fake} \tag{9}$$

where $I_{UL}$ and $\widehat{I_{UL}}$ represent the unlabeled image and its reconstructed version. $M_{UL}$ and $\widehat{M_{UL}}$ denote the mask predictions for the unlabeled image and the reconstructed unlabeled image, respectively.

Note that the gradients of $L_{Dice}$ in GRN-SEL and $L_{Customize}$ in GRN-SSL are passed through generator $G$ and segmentor $S$ simultaneously to enable mutual influence and enhance each other's performance during each training iteration.

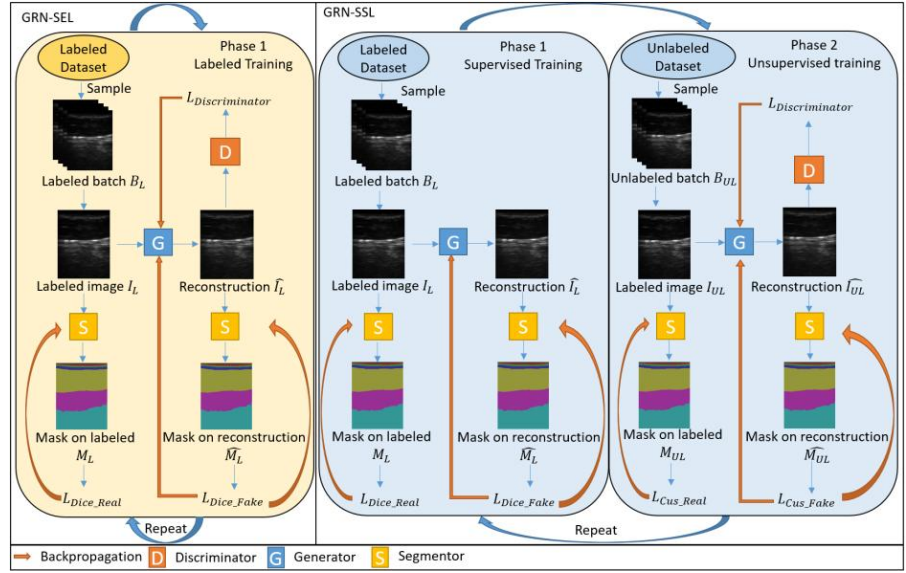

**Fig. 2.** Architectures of GRN-SEL and GRL-SSL. In GRN-SEL the generator receives feedback from the segmentor through backpropagation of the segmentation loss $L_{Dice\_Fake}$. GRN-SSL operates in two phases: Phase 1 (supervised training) and phase 2 (unsupervised training) with batches of equal size sampled from each dataset to ensure balanced contributions from labeled data and unlabeled data. The GRN-SSL

architecture is similar to that of GRN-SEL, except that during the unsupervised training phase, a customized loss that evaluates model prediction on unlabeled images is back-propagated to the generator.

---

**Algorithm 1** The Generative Reinforcement Network (GRN) Algorithm for sample-efficient learning

---

**Require:** $G_\theta$: Generator with trainable parameters $\theta$

**Require:** $D_{\theta'}$: Discriminator with trainable parameters $\theta'$

**Require:** $S_{\theta''}$: Segmentor with trainable parameters $\theta''$

**Require:** $Dataset_L(I, M)$: Collection of labeled samples

**Require:** $T$: total number of iterations

   *for t = 1, ..., T do*

      Sample $I_L, M \sim Dataset_L(I, M)$

      $\widehat{I_L} = G_\theta(I_L)$

      $P_L = D_{\theta'}(I_L)$

      $\widehat{P_L} = D_{\theta'}(\widehat{I_L})$

      $L_D = (L_{MSE}(P_L, 1) + L_{MSE}(\widehat{P_L}, 0))/2$ // 1:Real, 0: Fake

     **Update** $\theta'$ based on $L_D$

      $\widehat{M_L} = S_{\theta''}(\widehat{I_L})$

      $L_G = \lambda_{adv} L_{MSE}(\widehat{P_L}, 1) + \lambda_{seg} L_{Dice}(\widehat{M_L}, M_L) + \lambda_{L1} L1(\widehat{I_L}, I_L)$ // 1:Real, 0: Fake

     **Update** $\theta$ based on $L_G$

      $M_L = S_{\theta''}(I_L)$

      $L_S = (L_{Dice}(M_L, M), L_{Dice}(\widehat{M_L}, M))/2$

     **Update** $\theta''$ based on $L_S$

   **end for**

   **return** $\theta, \theta', \theta''$

---

**Algorithm 2** The Generative Reinforcement Network (GRN) Algorithm for semi-supervised learning

---

**Require:** $G_{\theta}$: Generator with trainable parameters $\theta$

**Require:** $D_{\theta'}$: Discriminator with trainable parameters $\theta'$

**Require:** $S_{\theta''}$: Segmentor with trainable parameters $\theta''$

**Require:** $Dataset_L(I, M)$: Collection of labeled samples

**Require:** $Dataset_{UL}(I, M)$: Collection of labeled samples

**Require:** $T$: total number of iterations

  **for** t = 1, ..., T do

    Sample $I_L, M \sim Dataset_L(I, M)$

$\widehat{I_L} = G_{\theta}(I_L)$

$\widehat{M_L} = S_{\theta''}(\widehat{I_L})$

$L_G = \lambda_{seg} L_{Dice}(\widehat{M_L}, M) + \lambda_{L1} L1(\widehat{I_L}, I_L)$

**Update** $\theta$ based on $L_G$

$M_L = S_{\theta''}(I_L)$

    $L_S = (L_{Dice}(M_L, M), L_{Dice}(\widehat{M_L}, M))/2$

**Update** $\theta''$ based on $L_S$

Sample $I_{UL} \sim Dataset_{UL}(I, M)$

$\widehat{I_{UL}} = G_{\theta}(I_{UL})$

$P_L = D_{\theta'}(I_{UL})$

$\widehat{P_L} = D_{\theta'}(\widehat{I_{UL}})$

    $L_D = (L_{MSE}(P_L, 1) + L_{MSE}(\widehat{P_L}, 0))/2$ // 1:Real, 0: Fake

    **Update** $\theta'$ based on $L_D$

$\widehat{M_{UL}} = S_{\theta''}(\widehat{I_{UL}})$

$L_G = \lambda_{adv} L_{MSE}(\widehat{P_L}, 1) + \lambda_{customize} L_{customize}(\widehat{M_{UL}}) + \lambda_{L1} L1(\widehat{I_{UL}}, I_{UL})$ // 1:Real, 0: Fake

**Update** $\theta$ based on $L_G$

$M_{UL} = S_{\theta''}(I_{UL})$

$L_S = (L_{customize}(\widehat{M_{UL}}) + L_{customize}(M_{UL}))/2$

**Update** $\theta''$ based on $L_S$

**end for**

**return** $\theta, \theta', \theta''$

### 2.2.2 Segmentation-guided Enhancement (SGE)

In the GRN framework, the objective function is defined as $min_G min_S L_{Seg}(S(G(x)), y)$, where the generator $G$ and the segmentor $S$ share the same goal to minimize the segmentor loss. Consequently, during the inference phase, the generator and the segmentor can be integrated to provide a joint prediction $y = S(G(x))$. This two-step process, termed Segmentation Guided Enhancement (**SGE**), ensures that image enhancements are directly designed to improve segmentation outcomes.

### 2.3 Interpolation Consistency Loss

Similar to a previous study [32], we adopt interpolation consistency loss as our customized loss for GRN-SSL. For each image in the unlabeled batch $B_{UL}$, we randomly sample a second image from the same batch. Both images are processed by the generator $G$ and segmentor $S$ to produce their respective segmentation masks. We then randomly generate an interpolation coefficient $\lambda$ from a predefined distribution to create a mixed image through an interpolation of the original and sampled images. This mixed image is subsequently processed by the generator and segmentor to obtain a corresponding segmentation mask. The interpolation consistency loss quantifies the alignment between the segmentation mask of the mixed image and the interpolated masks of the original and sampled images.

$$L_{ICT} = MSE(M_{mixed}, Mix_\lambda(M_{UL}, M_S)) \tag{11}$$

where $M_{mixed}$ denotes the predicted mask derived from the mixed image. $M_{UL}$ and $M_S$ are the mask predictions based on the unlabeled image and randomly sampled image, respectively. $Mix_\lambda()$ performs the interpolation of two images based on $Mix_\lambda(Tensor_a, Tensor_b) = \lambda Tensor_a + (1-\lambda)Tensor_b$.

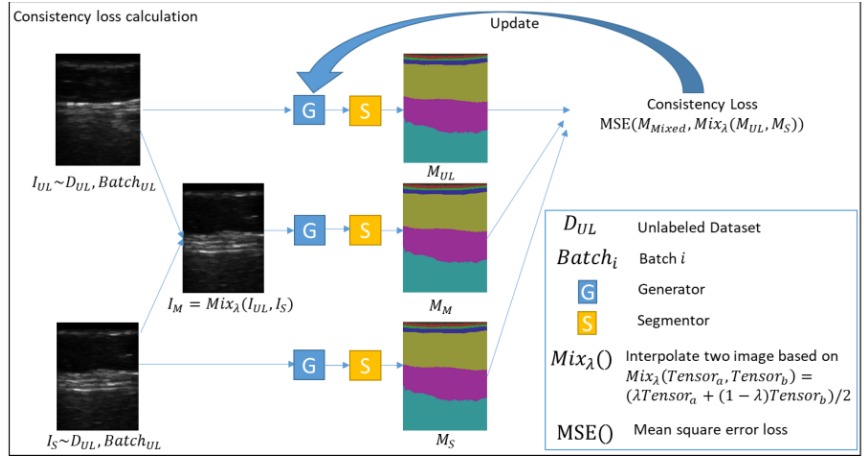

**Fig. 3** Interpolation Consistency Loss Calculation. $I_{UL}$ represents the unlabeled image, and $I_S$ denotes a randomly sampled image from the same batch. The interpolation consistency loss enforces the model to ensure that the predicted mask from the mixed image $I_M$ closely aligns with the interpolated mask derived from the unlabeled image $I_{UL}$ and the randomly sampled image $I_S$.

## 2.4 Backbone Network Architecture

The backbone segmentation network utilizes a 2D Unet model configured with encoder channels of (16, 32, 64, 128, 256). The model is designed to accept single-channel input and produce segmentation masks for seven classes, making it well-suited for our ultrasound imaging segmentation task. The generator encoder, illustrated in the upper section of Fig 5, incorporates residual blocks to improve gradient flow efficiency and facilitate the training of deeper network layers. The encoding process begins with an initial convolution layer that reduces spatial dimensions while increasing feature depth. This is followed by a series of residual blocks that downsample the input data. In contrast, the decoder utilizes 2D transposed convolutions to upscale high-dimensional feature representations, effectively reconstructing the images. Similar to the pix2pix model [33], we utilize a PatchGAN discriminator to assess the realism of generated images at the patch level [33]. The discriminator architecture comprises several convolution layers that progressively extract features from the input images. The discriminator outputs a patch-wise label map that evaluates the authenticity of each patch on the image.

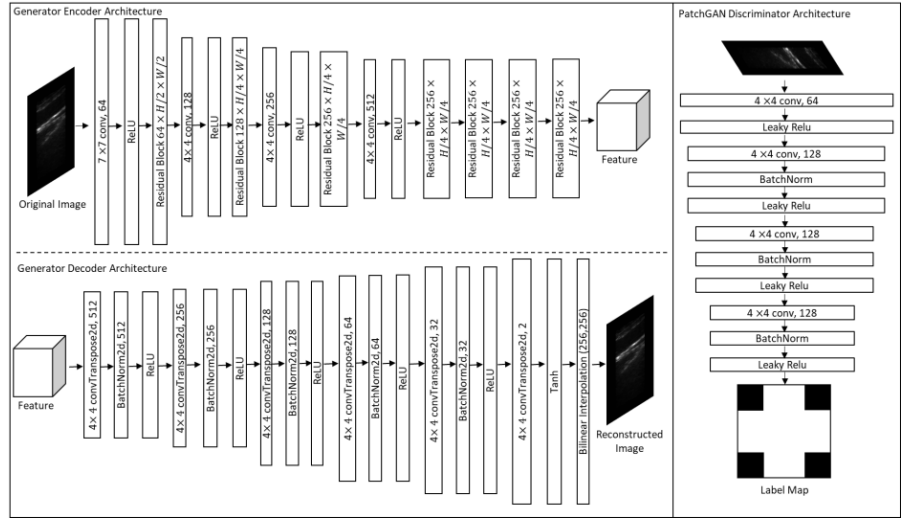

**Fig. 4.** Backbone network architecture. The encoder (upper section, left) utilizes multiple residual blocks to progressively capture high-dimensional patterns. The decoder (lower section, left) reconstructs the images from these high-level representations using transposed convolution layers, facilitating the accurate generation of segmentation masks.

### 2.5 Study Dataset

The primary dataset is **Lumbar Back Ultrasound Dataset (LBUD)**. It is acquired from an ongoing project funded by the National Institute of Health (NIH), which enrolls participants with cLBP and healthy controls (Institutional Review Board: STUDY22090014). All participants were positioned face-down on an examination table. The ultrasound array transducer was placed laterally to the midline at the lumbar 3-4 (L3-L4) vertebral interspace (Fig. 1, Part 1), specifically targeting the multifidus (MF) and erector spinae (ES) muscles. A row-column array (RCA) transducer (RC6gV, Vermon), operating at a center frequency of 6 MHz with an active aperture of 25.6 mm × 25.6 mm, was connected to the Vantage 256 ultrasound system (Verasonics Inc., WA, USA) to acquire 3-D volumetric ultrasound data. A synthetic aperture technique was used during acquisition to improve image quality. To acquire a single frame, each element of the transducer individually transmits one pulse in sequence until all the elements have performed transmissions, and after each transmission, all the elements simultaneously receive the echo signals. This enabled a synthetic aperture (SA) approach to achieve dynamic focusing during beamforming for image reconstruction.

Temporal compounding was applied by averaging three consecutive frames to improve the image signal-to-noise ratio. Both the right and left sides of the back were scanned. On each side, two specific locations over the MF and ES muscles were targeted by experienced ultrasound examiners (with 9 (ACB) and 25 (ADW) years of clinical experience). Each designated location was scanned three times, yielding a total of 12 B-mode scans per participant. A total of 29 participants were enrolled and used for this study. We randomly sampled 69 scans ($N_{image} = 17,664$) taken from various locations across 29 patients for algorithm development (Table I). An experienced ultrasound operator (Ms. Zhao) meticulously annotated six distinct anatomical layers depicted on the 3-D ultrasound scans, including the dermis, superficial fat, superficial fascial membrane (SFM), deep fat, deep fascial membrane (DFM), and MF muscle (Fig. 1, Parts 2 & 3). For the purpose of deep learning model training, the annotated B-mode images were divided into training ($N_{image} = 10,722, N_{scan} = 44, N_{patient} = 16$), internal validation ($N_{image} = 512, N_{scan} = 2, N_{patient} = 2$), and independent test ($N_{image} = 6,430, N_{scan} = 23, N_{patient} = 11$) sets at the patient level. To ensure a more reliable estimate of model performance, we set aside a relatively larger independent test set comprising 23 scans from different patients.

**Table 1.** Descriptive statistics information for demographic variables and LBP status.

| N = 29 | Summary |
|---|---|
| Demographic variable | |
| Age | 47.11 (19.34) |
| Female | 18 (62.07%) |
| Hispanic or Latino | 1 (3.44%) |
| Black or African American | 2 (6.90%) |
| White | 26 (89.65%) |
| Height (in inches) | 66.38 (3.80) |
| Weight(in pounds) | 179.83 (36.02) |

| | | |
|---|---|---|
| LBP status | | |
| | Positive | 21 (72.41%) |

Summary statistics are reported as mean (standard deviation) for continuous measurements and percentages for categorical measurements.

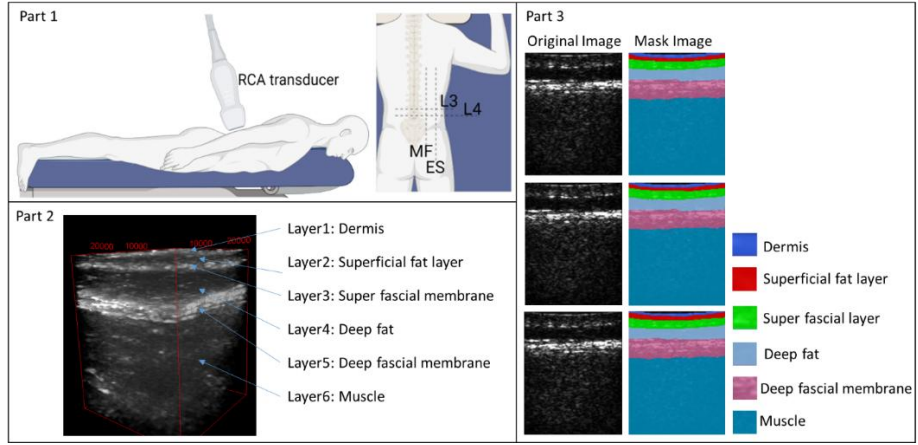

**Fig. 5.** 3-D ultrasound B-mode image acquisition and annotations. Part 1: ultrasound image acquisition. Part 2: A 3-D ultrasound B-mode image depicting six tissue layers. Part 3: Original B-mode images along the coronal or sagittal directions paired with their corresponding annotations, where each color represents a specific tissue layer.

To further assess the generalizability of the GRN model, we conducted state-of-the-art comparisons on two external datasets: **a dermoscopic skin lesion dataset** and **the Kvasir dataset**. The dermoscopic skin-lesion dataset contains 900 high-resolution dermoscopic RGB images (800 reserved for training and 100 for validation) [27], each paired with an expert-annotated binary mask indicating lesion regions, plus a public test set of 379 images. Sourced from clinics worldwide, it covers both benign nevi and malignant melanoma. All images resized to 256 × 256 pixels for uniformity.

Kvasir dataset provides 1000 colonoscopy frames that depicted colorectal polyps, each accompanied by a pixel-accurate ground-truth mask drawn and verified by gastroenterology experts. Image size ranges from 332 by 487 to 1920 by 1072. For our experiments, we randomly designated 800 images for training, 100 for validation, and 100 for testing.

**2.6 Model training and performance evaluation**

In line with standard practices in semi-supervised learning, we performed a state-of-the-art comparison of GRN methods against existing approaches on both the primary LBUD dataset and two external benchmark datasets (dermoscopic skin-lesion and Kvasir). Specifically, we designated 5%, 10%, and 20% of each training dataset as labeled subsets, with the remainder treated as unlabeled. We compared the performance of GRN variants (Table II) with that seven other semi-supervised learning methods, including MT[22], UAMT [23], Uncertainty Rectified Pyramid Consistency (URPC) [34], Regularized dropout (R-drop) [36], Interpolation Consistency Training (ICT) [32], and Deep Adversarial Network (DAN) [37]. All semi-supervised learning methods, including our GRN variants and six other comparison methods, utilized the UNet model as their backbone network.

For LBUD specifically, we further benchmarked GRN against fully supervised settings by allocating 30%, 40%, 50%, and 100% of the training dataset as labeled. This approach allows the evaluation of the effectiveness of GRN methods in reducing the need for extensive image annotation while maintaining and even enhancing segmentation performance. We further analyzed the effect of RAug, SGE, and SSL on model performance across varying proportions of labeled images for training. The effect of RAug was assessed by calculating the difference in dice similarity coefficient (DSC) between the **GRN-SEL** method and the fully supervised model. The effect of **SGE effect** was determined by comparing GRN variants with and without SGE, while the effect of **SSL** was evaluated by measuring the performance difference with or without the inclusion of unlabeled data for unsupervised training. Additionally, we conducted an ablation study on a 5% labeled dataset to evaluate the GRN model performance.

In training the GRN-SEL, $\lambda_{adv}$, $\lambda_{seg}$ and $\lambda_{L1}$ in $L_G$ are set to 1, 100, 100, respectively. For GRN-SSL training, $\lambda_{seg}$ and $\lambda_{L1}$ are both set to 100. In the unsupervised training of GRN-SSL, $\lambda_{adv}$, $\lambda_{customize}$ and $\lambda_{L1}$ are set to be 1, 1, 100, respectively. Following the approach of the pix2pix model [33], a smaller weight was assigned to the adversarial loss. The Adam optimizer was used for training the generator, segmentor, and discriminator, with an initial learning rate of 0.0002 and exponential decay rates $\beta$ set to be (0.5, 0.999). The maximum number of epochs was set to 50. The model with the lowest dice loss on the validation dataset was selected. An early stopping mechanism was implemented, terminating the training if validation performance did not improve for five consecutive epochs. The primary evaluation metrics of the segmentation model were DSC, with a higher DSC indicating a greater overlap between the predicted mask and the ground truth mask. In addition, three supplementary metrics, including average surface distance (ASD), Hausdorff Distance 95% (HD95), and Intersection over Union (IoU), were also calculated. In this study, all methods were implemented using PyTorch = 2.3.1 on an NVIDIA GTX3090 GPU.

## 3. Results
### 3.1 Performance evaluations and comparison on LBUD Dataset

Table 2 presents the segmentation performance of our proposed GRN methods compared to existing techniques on the independent test set with varying proportions of labeled data (5%, 10%, 20%) involved in the training. Across all labeled data proportions, the GRN-SSL variant and GRN-SEL with SGE consistently outperformed other methods (p-value < 0.05). When trained on 5% labeled data, GRN-SEL achieved superior results compared to 5 out of 7 models without utilizing unlabeled images. At the 10% labeled level, GRN-SSL (w SGE) is only 3.46% lower than the fully supervised model trained on all labeled data At the 20% labeled level, GRN-SSL with SGE achieved an 80.05% DSC, which is just 0.57% lower than the fully supervised model (80.62%); however, this difference is still statistically significant (p < 0.05).

Table 3 summarizes the mean DSC for each tissue layer as well as the overall mean when trained on 5% of the labeled images in the training set. The GRN-SSL variant with SGE outperformed all other methods in 3 out of 6 tissue layers significantly, while GRN-SSL exceeded the performance of other methods in 2 out of 6 layers. Although the overall DSC for GRN-SEL with SGE was higher, it did not achieve a statistically significant improvement in the segmentation of individual layers. Some examples in Fig. 6 show the prediction results of the GRN methods and other models.

**Table 2**. Comparison of segmentation performance between our methods and existing methods on LBUD dataset in terms of 95% Hausdorff Distance (95HD), Average Surface Distance (ASD), Intersection over Union (IOU), and DSC (95% confidence interval).

| Labeling level | Methods | HD95 | ASD | IOU | DSC |
|---|---|---|---|---|---|
| 5% | Fully supervised | 54.42 (54.04, 54.80) | 19.54 (19.36, 19.72) | 34.16 (33.90, 34.43) | 43.97 (43.71, 44.24) |
| | MT | 51.83 (51.49, 52.16) | 15.81 (15.70, 15.93) | 50.99 (50.68, 51.31) | 65.09 (64.78, 65.40) |
| | UAMT | 42.29 (41.84, 42.73) | 12.15 (11.92, 12.31) | 50.78 (50.44, 51.22) | 64.39 (64.04, 64.74) |
| | ICT | 35.18 (34.87, 35.50) | 11.16 (11.05, 11.28) | 52.15 (51.83, 52.47) | 66.07 (65.76, 66.39) |
| | R-drop | 50.28 (49.87, 50.69) | 13.71 (13.58, 13.85) | 52.11 (51.78, 52.44) | 65.92 (65.59, 66.25) |
| | URPC | 44.47 (43.77, 45.16) | 14.64 (14.33, 14.95) | 45.28 (44.84, 45.72) | 56.98 (56.49, 57.47) |
| | DAN | 38.31 (37.94, 38.65) | 11.04 (10.93, 11.15) | 50.39 (50.11, 50.67) | 64.56 (64.30, 64.82) |
| | GRN-SEL | 38.48 (38.12, 38.83) | 11.45 (11.33, 11.58) | 51.35 (51.01, 51.70) | 65.16 (64.81, 65.50) |
| | GRN-SEL (SGE) | **33.82 (33.41, 34.23)**[*] | **9.94 (9.81, 10.08)**[*] | 53.17 (52.81, 53.53)[*] | 66.55 (66.20, 66.90)[*] |
| | GRN-SSL | 49.21 (48.87, 49.56) | 15.80 (15.67, 15.93) | **53.45 (53.14, 53.77)**[*] | **67.33 (67.03, 67.63)**[*] |
| | GRN-SSL (SGE) | 45.35 (44.99, 45.72) | 14.47 (14.35, 14.59) | **55.31 (55.01, 55.61)**[*] | **68.90 (68.61, 69.19)**[*] |
| 10% | Fully supervised | 29.72 (29.32, 30.12) | 8.76 (8.61, 8.91) | 52.67 (52.37, 52.97) | 64.87 (64.57, 65.18) |
| | MT | 30.94 (30.57, 31.30) | 9.00 (8.86, 9.13) | 60.00 (59.68, 60.33) | 77.35 (75.20, 79.49) |
| | UAMT | 30.17 (29.74, 30.59) | 8.90 (8.75, 9.06) | 60.11 (59.77, 60.46) | 72.75 (72.42, 73.09) |
| | ICT | 28.83 (28.50, 29.15) | 8.26 (8.15, 8.38) | 61.30 (61.02, 61.58) | 74.06 (73.81, 74.31) |
| | R-drop | 33.76 (32.43, 33.09) | 8.97 (8.86, 9.08) | 62.97 (62.69, 63.25) | 75.31 (75.06, 75.57) |

|  | Method | | | | |
|---|---|---|---|---|---|
|  | URPC | 28.10 (27.68, 28.52) | 7.84 (7.70, 7.99) | 56.71 (56.40, 57.02) | 69.34 (69.02, 69.65) |
|  | DAN | 32.74 (32.40, 33.09) | 9.53 (9.41, 9.66) | 60.00 (59.71, 60.28) | 73.00 (72.73, 73.27) |
|  | GRN-SEL | 30.94 (30.58, 31.30) | 9.19 (9.07, 9.31) | 60.15 (59.85, 60.45) | 73.08 (72.80, 73.36) |
|  | GRN-SEL (SGE) | **22.55 (22.24, 22.85)**[*] | **6.42 (6.32, 6.51)**[*] | **65.15 (64.86, 65.44)**[*] | **76.98 (76.72, 77.23)**[*] |
|  | GRN-SSL | 31.18 (30.83, 31.54) | 8.68 (8.56, 8.81) | **63.71 (63.41, 54.01)**[*] | **75.65 (75.37, 75.93)**[*] |
|  | GRN-SSL (SGE) | 29.25 (28.94, 29.57) | **7.69 (7.59, 7.80)**[*] | **65.74 (65.42, 66.05)**[*] | **77.16 (76.87, 77.44)**[*] |
| 20% | Fully supervised | 25.66 (25.28, 26.03) | 7.68 (7.55, 7.81) | 62.34 (62.03, 62.66) | 74.52 (74.23, 74.81) |
|  | MT | 22.79 (22.40, 23.18) | 6.25 (6.14, 6.37) | 64.96 (64.69, 65.24) | 76.82 (76.57, 77.07) |
|  | UAMT | 22.99 (22.60, 23.38) | 6.21 (6.10, 6.32) | 65.15 (64.87, 65.44) | 76.95 (76.69, 77.21) |
|  | ICT | 26.28 (25.88, 26.68) | 7.05 (6.95, 7.15) | 64.50 (64.25, 64.75) | 76.54 (76.31, 77.21) |
|  | R-drop | 21.45 (21.06, 21.84) | 5.72 (5.61, 5.83) | 65.70 (65.41, 65.99) | 77.35 (77.09, 77.61) |
|  | URPC | 25.51 (24.93, 26.09) | 6.72 (6.57, 6.87) | 62.05 (61.73, 62.37) | 73.82 (73.50, 74.13) |
|  | DAN | 30.62 (30.25, 30.98) | 8.74 (8.63, 8.85) | 61.76 (61.48, 62.05) | 74.34 (74.08, 74.60) |
|  | GRN-SEL | 27.23 (26.84, 27.63) | 7.89 (7.75, 8.03) | 63.49 (63.20, 63.79) | 75.72 (75.45, 75.99) |
|  | GRN-SEL (SGE) | **18.44 (18.10, 18.78)**[*] | **5.37 (5.26, 5.48)**[*] | **67.61 (67.33, 67.89)**[*] | **78.85 (78.60, 79.10)**[*] |
|  | GRN-SSL | **17.52 (17.19, 17.84)**[*] | **4.85 (4.76, 4.94)**[*] | **66.02 (65.77, 66.27)**[*] | **77.61 (77.39, 77.84)**[*] |
|  | GRN-SSL (SGE) | **14.63 (14.30, 14.95)**[*] | **4.13 (4.05, 4.22)**[*] | **69.07 (68.81, 69.33)**[*] | **80.05 (79.83, 80.27)**[*] |
| 100% | Fully supervised | 12.02 (11.82, 12.22) | 3.28 (3.24, 3.33) | 69.91 (69.74, 70.08) | 80.62 (80.47, 80.77) |

Bold denotes the proposed methods that outperform all other SSL methods. * denotes statistically significant outperforms all other SSL methods ($p < 0.05$, paired t-test).

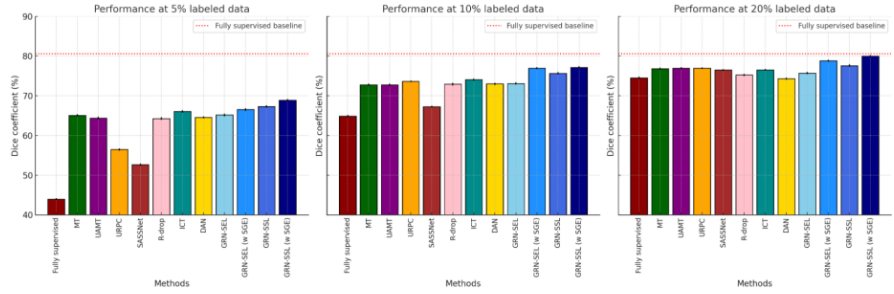

**Fig. 6**. Comparison of segmentation model performance across various methods trained on different amounts of labeled data (5%, 10%, and 20%). The red dotted line indicates the performance of the model when trained on 100% labeled images in the training set. The small bars on each column indicate the confidence intervals (CI) for the segmentation results of each method.

**Table 3.** Comparison of segmentation performance for each tissue layer among GRN methods and other models trained on the 5% labeled images in the LBUD dataset, expressed in terms of DSC (95% confidence interval).

| Methods | Dermis | Superficial Fat Layer | SFM | Deep Fat | DFM | Muscle | Overall |
|---|---|---|---|---|---|---|---|
| Fully supervised | 7.06 (6.87, 7.25) | 0.67 (0.64, 0.70) | 57.69 (57.39, 57.99) | 63.76 (63.01, 64.51) | 65.78 (65.34, 66.22) | 68.89 (68.57, 69.21) | 43.97 (43.71, 44.23) |
| MT | 62.98 (62.66, 63.30) | 59.81 (59.40, 60.22) | 69.45 (69.06, 69.84) | 62.41 (61.67, 63.15) | 66.40 (65.98, 66.82) | 69.49 (69.14, 69.84) | 65.09 (64.78, 65.40) |
| UAMT | 60.18 (59.80, 60.56) | 52.43 (51.93, 52.93) | 71.09 (70.63, 71.55) | 63.06 (62.31, 63.81) | 69.14 (68.71, 69.57) | 70.44 (70.10, 70.78) | 64.39 (64.04, 64.74) |
| URPC | 46.78 (46.14, 47.42) | 18.71 (18.33, 19.09) | 64.96 (64.35, 65.57) | 66.90 (66.11, 67.69) | 67.03 (66.59, 67.47) | 74.36 (73.97, 74.75) | 56.46 (56.17, 56.75) |
| SASSNet | 34.33 (33.69, 34.97) | 27.29 (26.68, 27.90) | 58.28 (57.75, 58.81) | 68.30 (67.54, 69.06) | 60.13 (59.65, 60.61) | 67.62 (67.31, 67.93) | 52.66 (52.38, 52.94) |
| R-drop | 65.05 (64.75, 65.35) | 55.16 (54.65, 55.67) | 68.70 (68.30, 69.10) | 67.18 (66.46, 67.90) | 66.65 (66.22, 67.08) | 72.32 (72.02, 72.62) | 65.92 (65.59, 66.25) |
| ICT | 65.91 (65.63, 66.19) | 55.44 (54.97, 55.91) | 70.25 (69.83, 70.67) | 64.40 (63.69, 65.11) | 68.68 (68.26, 69.10) | 71.75 (71.47, 72.03) | 66.07 (65.75, 66.39) |
| DAN | 62.59 (62.30, 62.88) | 46.79 (46.45, 47.13) | 73.93 (73.68, 74.18) | 63.77 (63.09, 64.45) | 67.95 (67.50, 68.40) | 72.35 (72.08, 72.62) | 64.56 (64.30, 64.82) |
| GRN-SEL | 63.58 (63.20, 63.96) | 54.87 (54.39, 55.35) | 69.83 (69.44, 70.22) | 63.90 (63.17, 64.63) | 68.28 (67.85, 68.71) | 70.47 (70.04, 70.90) | 65.16 (64.82, 65.50) |
| GRN-SEL (w SGE) | 65.15 (64.76, 65.54) | 56.73 (56.25, 57.21) | 71.23 (70.82, 71.64) | 65.83 (65.06, 66.60) | 68.61 (68.16, 69.06) | 71.70 (71.37, 72.03) | **66.55 (66.21, 66.89)**[*] |
| GRN-SSL | **67.06 (66.77, 67.35)**[*] | 58.88 (58.46, 59.30) | 68.85 (68.41, 69.29) | 67.46 (66.76, 68.16) | 67.14 (66.73, 67.55) | **74.57 (74.16, 74.98)**[*] | **67.33 (67.03, 67.63)**[*] |
| GRN-SSL (w SGE) | **69.40 (69.11, 69.69)**[*] | **61.91 (61.53, 62.29)**[*] | 72.11 (71.69, 72.53) | 68.15 (67.44, 68.86) | 67.17 (66.80, 67.54) | **74.63 (74.35, 74.91)**[*] | **68.90 (68.62, 69.18)**[*] |

Bold denotes the proposed methods that outperform all other SSL methods. * denotes statistically significant outperforms all other SSL methods ($p < 0.05$, paired t-test).

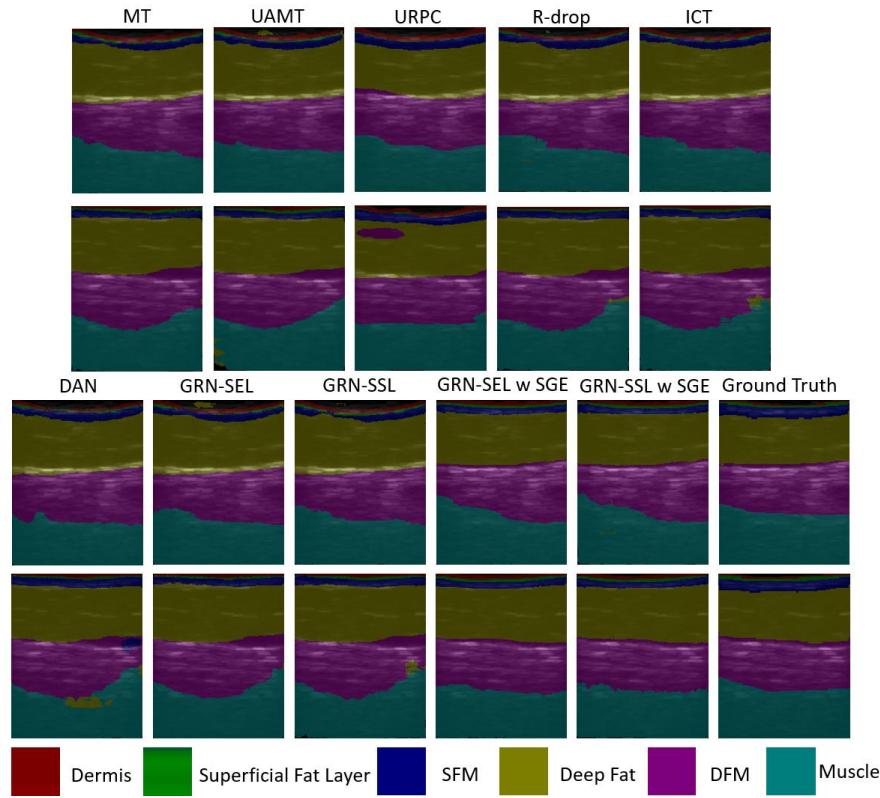

**Fig. 7**. Mask predictions for GRN methods and other methods when trained on the 5% labeled images in the training set. GRN variants, particularly those utilizing SGE, demonstrate superior segmentation performance and higher accuracy in tissue layer segmentation.

### 3.2 Segmentation performance across varying proportions of labeled data

Table 4 presents the segmentation performance of GRN variants on varying proportions of labeled images in the LBUD training set. The results indicate that with just 30% of the labeled images, the GRN-SEL with SGE achieved performance comparable to the model trained on 100% of the labeled images. Additionally, when utilizing 30% of the labeled dataset, the GRN-SSL with SGE attained performance statistically similar to the model trained on all labeled images. As the labeled image proportion increased to 40%, the GRN-SSL model matched the performance of the fully supervised approach. Even without utilizing the unlabeled dataset, the GRN-SEL achieved performance similar to those of the supervised model. Notably, the

GRN framework not only maintained equivalence with fully supervised models under reduced labeling conditions but also exceeded their performance limits. Specifically, the GRN-SSL with SGE achieved a DSC that was 2.59% higher than that of the fully supervised model while utilizing only 40% of the labeled images in the training set. These findings suggest the efficacy of the GRN methods in enhancing segmentation performance while significantly reducing the need for extensive data labeling.

**Table 4**. Segmentation performance of the proposed GRN methods on the independent test set when training on varying proportions of labeled images in the training set, expressed in terms of DSC (Standard deviation).

| Method | 5% | 10% | 20% | 30% | 40% | 50% | 100% |
|---|---|---|---|---|---|---|---|
| Fully supervised | 43.97 (43.71, 44.23) | 64.87 (64.57, 65.17) | 74.52 (74.23, 74.81) | 80.21 (80.04, 80.38) | 80.53 (80.36, 80.70) | 80.49 (80.33, 80.65) | 80.62 (80.47, 80.77) |
| GRN-SEL | 65.16 (64.82, 65.50) | 73.08 (72.80, 73.36) | 75.72 (75.45, 75.99) | 79.55 (79.38, 79.72) | **80.79 (80.64, 80.94)**[*] | 80.04 (79.90, 80.18) | 80.53 (80.39, 80.67) |
| GRN-SEL (w. SGE) | 66.55 (66.21, 66.89) | 76.98 (76.72, 77.24) | 78.85 (78.60, 79.10) | **82.60 (82.45, 82.75)**[*] | **83.04 (82.90, 83.18)**[*] | **82.18 (82.04, 82.32)**[*] | **82.45 (82.33, 82.57)**[*] |
| GRN-SSL | 67.33 (67.03, 67.63) | 75.65 (75.37, 75.93) | 77.61 (77.27, 77.95) | 79.90 (79.73, 80.07) | **80.85 (80.70, 81.00)**[*] | 80.53 (80.39, 80.67) | - |
| GRN-SSL (w. SGE) | 68.90 (68.62, 69.18) | 77.16 (76.86, 77.46) | 80.05 (79.81, 80.29) | **81.86 (81.70, 82.02)**[*] | **83.21 (83.07, 83.35)**[*] | **82.45 (82.32, 82.58)**[*] | - |

Bold denotes the proposed methods that outperform all other SSL methods. * denotes statistically significant outperforms all other SSL methods ($p < 0.05$, paired t-test).

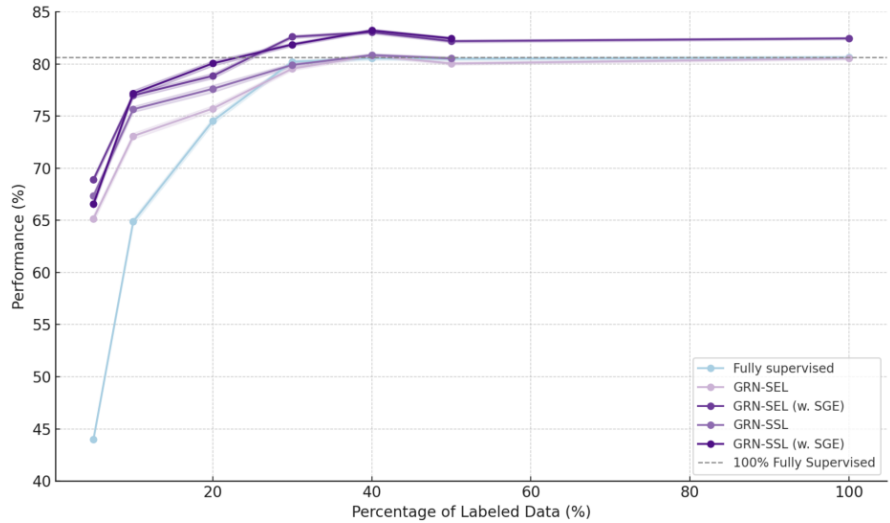

**Fig. 8.** Segmentation performance of GRN variants relative to the proportion of the labeled images in the training set. The dotted line indicates the performance of the models trained on all labeled images.

### 3.3 GRN Component Evaluation

Table 5 presents the individual and combined effects of DGA, SGE, and SSL on model performance across different proportions (5%, 10%, 20%, 30%, 40%, 50%, and 100%) of labeled images in the LBUD training set. The results indicate that DGA has a substantial positive impact on model performance, particularly when the labeled data is scarce. Specifically, at lower labeling proportions (5% and 10%), DGA contributes to substantial performance gains of +21.19% and +8.21%, respectively. However, its positive impact diminishes as the number of labeled images increases, with the effect dropping to -0.09% when all labeled images are used. In contrast, SGE consistently enhanced model performance across all proportions of labeled images, showing stable and robust improvements. The inclusion of SGE leads to positive effects ranging from +1.13% at 20% labeled images to +3.89% at 10%.

**Table 5.** Effect of DGA, SGE, and SSL on segmentation performance gains across various proportions of labeled images in the training set, expressed in terms of DSC (Standard deviation).

| Variable | 5% | 10% | 20% | 30% | 40% | 50% | 100% |
|---|---|---|---|---|---|---|---|
| RAug | +21.19 (+20.76, +21.62) | +8.21 (+8.09, +8.33) | +0.89 (+0.72, +1.06) | +0.38 (+0.18, +0.58) | +0.36 (+0.22, +0.50) | **-0.46 (-0.53, -0.39)** | **-0.09 (-0.17, -0.01)** |
| SGE w. SEL | +1.39 (+1.32, +1.46) | +3.89 (+3.81, +3.97) | +1.13 (+0.96, +1.30) | +1.41 (+1.26, +1.56) | +1.32 (+1.17, +1.47) | +2.14 (+2.08, +2.20) | +1.91 (+1.86, +1.96) |
| SGE w. SSL | +1.57 (+1.52, +1.62) | +1.51 (+1.45, +1.57) | +1.21 (+1.08, +1.34) | +1.40 (+1.28, +1.52) | +1.39 (+1.15, +1.63) | +1.91 (+1.86, +1.96) | |
| SSL w. no SGE | +2.17 (+2.07, +2.27) | +2.57 (+2.47, +2.67) | +0.30 (+0.17, +0.43) | +0.69 (+0.59, +0.79) | +1.01 (+0.88, +1.14) | +0.49 (+0.44, +0.54) | |
| SSL w. SGE | +2.36 (+2.24, +2.48) | +0.18 (+0.09, +0.27) | +0.39 (+0.27, +0.51) | +0.68 (+0.58, +0.78) | +1.08 (+0.96, +1.20) | +0.26 (+0.21, +0.31) | |

Bold text denotes the proposed methods that statistically decrease the model performance. ($CI_{bounds}$ < 0, p < 0.05, paired t-test).

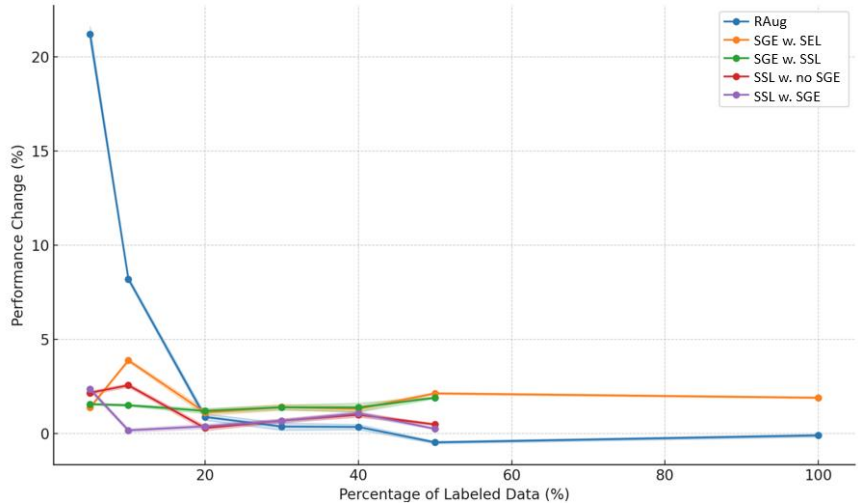

**Fig. 9**. Model performance with different GRN components across varying percentages of labeled images. The Raug significantly enhances performance at lower percentages of data images but shows a diminishing effect as more labeled data is incorporated. In contrast, SSL consistently shows a strong positive impact across most data proportions.

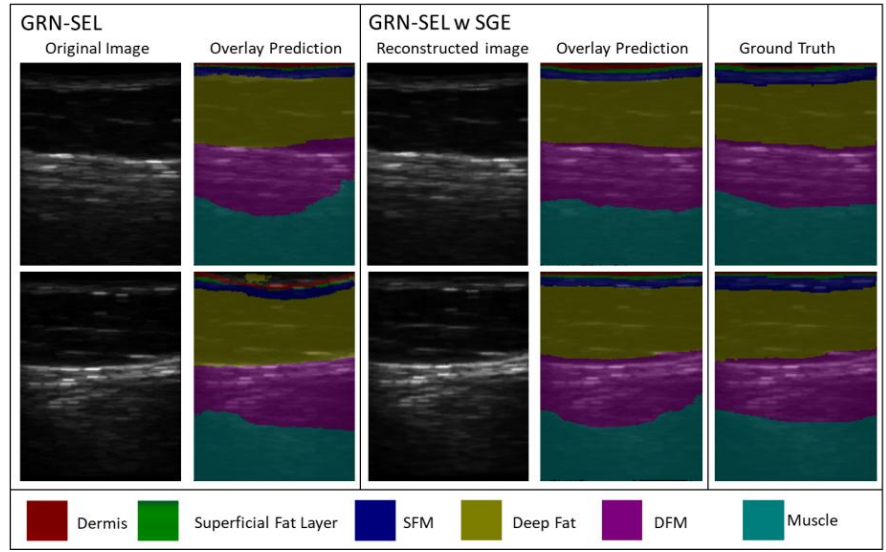

**Fig. 10**. Comparison of non-enhanced and enhanced images with corresponding mask prediction. The segmentation models achieve higher accuracy when utilizing SGE.

### 3.4 External Dataset Experiment

Table 6 presents the segmentation performance of various methods on the demoscopic skin-lesion dataset. GRN-SSL demonstrates superior segmentation performance, outperforming the other methods across all three labeling levels in terms of DSC. At the 10% and 20% labeling levels, its advantage over other SSL methods is marginally significant (p-value < 0.10, paired t-test). Specifically, when trained on a 20% labeling level, GRN-SSL achieved the highest DSC of 80.70%, which is a 4.64% improvement compared to the fully supervised model trained without utilizing any semi-supervised learning methods. Fig. 8 visualizes the mask prediction of GRN-SSL and other SSL methods.

**Table 6**. Comparison of segmentation performance between our methods and existing methods on dermoscopic skin-lesion dataset in terms of 95% Hausdorff Distance (95HD), Average Surface Distance (ASD), Intersection over Union (IOU), and DSC (95% confidence interval).

| Labeling level | Methods | HD95 | ASD | IOU | DSC |
|---|---|---|---|---|---|
| 5% | Fully supervised | 95.11 (92.15, 98.06) | 44.52 (42.03, 47.02) | 54.65 (52.02, 57.28) | 66.54 (64.03, 69.04) |
| | MT | 67.50 (62.71, 72.28) | 34.28 (31.19, 37.38) | 61.45 (58.70, 64.19) | 72.02 (69.53, 74.50) |
| | UAMT | 64.95 (60.16, 69.74) | 31.01 (28.20, 33.82) | 62.88 (60.18, 65.59) | 73.33 (70.87, 75.73) |
| | ICT | 64.22 (59.79, 68.65) | 31.28 (28.62, 33.94) | 61.89 (59.31, 64.48) | 72.86 (70.54, 75.19) |
| | R-drop | 73.00 (68.35, 77.66) | 35.90 (32.98, 38.81) | 61.46 (58.74, 77.66) | 72.09 (69.62, 74.55) |
| | URPC | 62.34 (57.00, 67.68) | 25.15 (22.35, 27.95) | 62.37 (59.63, 65.10) | 72.78 (70.32, 75.25) |
| | DAN | 56.83 (52.05, 61.61) | 23.01 (20.40, 25.61) | 62.97 (60.25, 65.68) | 73.30 (70.85, 75.76) |
| | GRN-SSL | 60.58 (56.23, 64.93) | 27.67 (25.29, 30.04) | 62.90 (60.33, 65.47) | **73.69 (71.38, 76.00)** |
| 10% | Fully supervised | 72.13 (67.76, 76.51) | 32.86 (30.37, 35.36) | 63.30 (60.79, 65.82) | 74.19 (71.96, 76.43) |
| | MT | 52.72 (48.28, 57.15) | 23.75 (21.33, 26.17) | 67.23 (64.77, 69.69) | 77.35 (75.20, 79.49) |
| | UAMT | 61.29 (56.72, 65.86) | 26.96 (24.47, 29.45) | 66.58 (64.14, 69.01) | 76.97 (74.87, 79.06) |
| | ICT | 53.86 (49.54, 58.17) | 24.80 (22.31, 27.29) | 67.65 (65.22, 70.07) | 77.76 (75.66, 79.86) |
| | R-drop | 56.09 (51.76, 60.43) | 25.35 (22.90, 27.80) | 66.81 (64.37, 69.26) | 77.07 (74.93, 79.21) |
| | URPC | 46.23 (41.54, 50.91) | 16.00 (13.85, 18.14) | 67.87 (65.41, 70.32) | 77.86 (75.75, 79.98) |
| | DAN | 45.62 (41.12, 50.14) | 15.68 (13.55, 17.82) | 66.90 (64.51, 69.29) | 77.27 (75.19, 79.35) |
| | GRN-SSL | 54.13 (49.63, 58.63) | 24.53 (22.02, 27.05) | **68.40 (65.96, 70.84)**[*] | **78.23 (76.10, 80.37)**[†] |
| 20% | Fully supervised | 67.40 (62.71, 72.08) | 30.66 (27.90, 33.41) | 66.19 (63.58, 68.80) | 76.06 (73.70, 78.41) |
| | MT | 55.42 (50.77, 60.06) | 25.50 (22.85, 28.15) | 69.53 (67.00, 72.06) | 78.81 (76.59, 81.03) |
| | UAMT | 48.23 (43.78, 52.68) | 21.39 (19.00, 23.78) | 71.14 (68.67, 73.61) | 80.14 (78.00, 82.28) |
| | ICT | 52.62 (48.15, 57.09) | 23.53 (21.02, 26.05) | 70.19 (67.72, 72.67) | 79.43 (77.27, 81.60) |
| | R-drop | 56.97 (52.39, 61.55) | 25.33 (22.79, 27.86) | 68.46 (66.01, 70.90) | 78.25 (76.10, 80.40) |
| | URPC | 46.23 (41.54, 50.91) | 16.00 (13.85, 18.14) | 67.87 (65.41, 70.32) | 77.86 (75.75, 79.98) |
| | DAN | 45.62 (41.12, 50.14) | 15.68 (13.55, 17.82) | 66.90 (64.51, 69.29) | 77.27 (75.19, 79.35) |
| | GRN-SSL | 46.66 (42.22, 51.09) | 20.16 (17.78, 22.55) | **71.80 (69.37, 74.23)**† | **80.70 (78.60, 82.80)**† |
| 100% | Fully supervised | 34.31 (30.63, 38.00) | 12.60 (11.14, 14.06) | 79.38 (77.77, 80.99) | 87.42 (86.18, 88.67) |

Bold text denotes the proposed methods outperform all other SSL methods. * denotes improvement is statistically significant (p < 0.05, paired t-test). † denotes improvement is marginal significant (p < 0.10, paired t-test).

Table 7 compares GRN-SSL with other SSL approaches on the Kvasir dataset across 5%, 10% and 20% labeling level. GRN-SSL delivers the best HD95, IOU, and DSC at every labeling level. At 20% labels, it is statistically superior to all SSL baselines on HD95, IOU, DSC, and ASD, and raises DSC by 16.63% compared to fully supervised model. Fig. 9 shows overlays of predicted masks on the original images for each SSL method.

**Table 7**. Comparison of segmentation performance between our methods and existing methods on kvasir dataset in terms of 95% Hausdorff Distance (95HD), Average Surface Distance (ASD), Intersection over Union (IOU), and DSC (95% confidence interval).

| Labeling level | Methods | HD95 | ASD | IOU | DSC |
|---|---|---|---|---|---|
| 5% | Fully supervised | 103.70 (96.47, 111.76) | 48.37 (44.06, 53.09) | 30.43 (26.42, 34.53) | 42.81 (38.11, 47.52) |
| | MT | 137.25 (130.88, 143.85) | 67.01 (63.14, 71.36) | 15.55 (13.28, 17.81) | 25.31 (22.19, 28.43) |
| | UAMT | 100.09 (92.66, 108.00) | 47.51 (43.03, 52.27) | 30.80 (26.69, 35.10) | 43.00 (38.26, 47.87) |
| | ICT | 111.01 (103.95, 118.68) | 55.19 (50.60, 60.15) | 28.95 (25.12, 32.91) | 41.27 (36.78, 45.80) |
| | R-drop | 134.67 (128.11, 141.79) | 69.64 (64.94, 74.22) | 21.10 (17.90, 24.47) | 32.10 (28.19, 36.08) |
| | URPC | 121.43 (114.76, 128.10) | 52.29 (48.06, 56.51) | 20.41 (17.59, 23.22) | 31.73 (28.10, 35.35) |
| | DAN | 115.10 (107.90, 122.30) | 49.48 (44.82, 54.51) | 23.65 (20.32, 26.99) | 35.47 (31.42, 39.52) |
| | GRN-SSL | **98.76 (91.95, 105.94)** | **46.52 (42.61, 50.91)** | 30.96 (27.31, 34.64) | 44.03 (39.76, 48.31) |
| 10% | Fully supervised | 98.82 (91.74, 106.22) | 46.25 (41.91, 50.86) | 31.73 (27.95, 35.85) | 44.38 (39.76, 49.02) |
| | MT | 118.31 (111.83, 125.26) | 57.59 (53.43, 62.19) | 26.77 (23.24, 30.42) | 39.04 (34.86, 43.29) |
| | UAMT | 100.27 (93.32, 108.13) | 45.39 (41.59, 49.52) | 27.63 (23.90, 31.58) | 39.74 (35.13, 44.36) |
| | ICT | 117.00 (110.19, 124.30) | 57.67 (53.51, 62.33) | 26.02 (22.44, 29.81) | 37.97 (33.74, 42.31) |
| | R-drop | 149.41 (143.14, 156.05) | 79.88 (74.67, 85.41) | 14.66 (12.41, 16.86) | 23.99 (20.89, 27.13) |
| | URPC | 130.60 (124.12, 137.08) | 66.03 (61.56, 70.49) | 19.33 (16.42, 22.25) | 30.10 (26.41, 33.79) |
| | DAN | 144.40 (137.66, 151.14) | 65.28 (59.34, 71.21) | 14.68 (12.40, 16.95) | 24.01 (20.88, 27.14) |
| | GRN-SSL | **97.63 (90.26, 105.59)** | 46.50 (41.72, 51.63) | **32.33 (28.25, 36.54)**[*] | **44.89 (40.02, 49.72)**[*] |
| 20% | Fully supervised | 100.22 (92.25, 108.54) | 46.62 (42.08, 51.34) | 32.54 (28.49, 36.68) | 45.34 (40.65, 49.97) |
| | MT | 114.18 (107.22, 121.80) | 56.08 (51.79, 60.70) | 29.40 (25.40, 33.44) | 41.78 (37.18, 46.30) |
| | UAMT | 85.78 (78.11, 93.53) | 38.25 (33.58, 43.28) | 36.44 (32.28, 41.08) | 49.25 (44.20, 54,15) |
| | ICT | 106.89 (99.65, 114.77) | 50.47 (45.86, 55.09) | 31.49 (27.36, 35.55) | 44.15 (39.53, 48.76) |

|      | Method | | | | |
|------|--------|---|---|---|---|
|      | R-drop | 121.35 (115.14, 128.01) | 58.50 (54.91, 62.38) | 19.06 (16.29, 21.91) | 29.86 (26.31, 33.45) |
|      | URPC | 115.11 (107.70, 122.52) | 46.78 (42.47, 51.09) | 29.65 (25.78, 33.52) | 42.33 (37.83, 46.82) |
|      | DAN | 144.42 (137.85, 150.99) | 75.99 (71.25, 80.73) | 17.24 (14.56, 19.93) | 27.37 (23.88, 30.86) |
|      | GRN-SSL | **64.81 (55.64, 72.91)**[*] | **23.70 (19.50, 28.50)**[*] | **49.60 (44.57, 54.42)**[*] | **61.96 (56.96, 66.65)**[*] |
| 100% | Fully supervised | 51.03 (43.25, 58.93) | 18.30 (14.66, 22.09) | 62.94 (57.68, 68.09) | 73.51 (68.89, 78.08) |

Bold text denotes the proposed methods statistically not inferior to 100% fully supervised model. ($p < 0.05$, paired t-test).

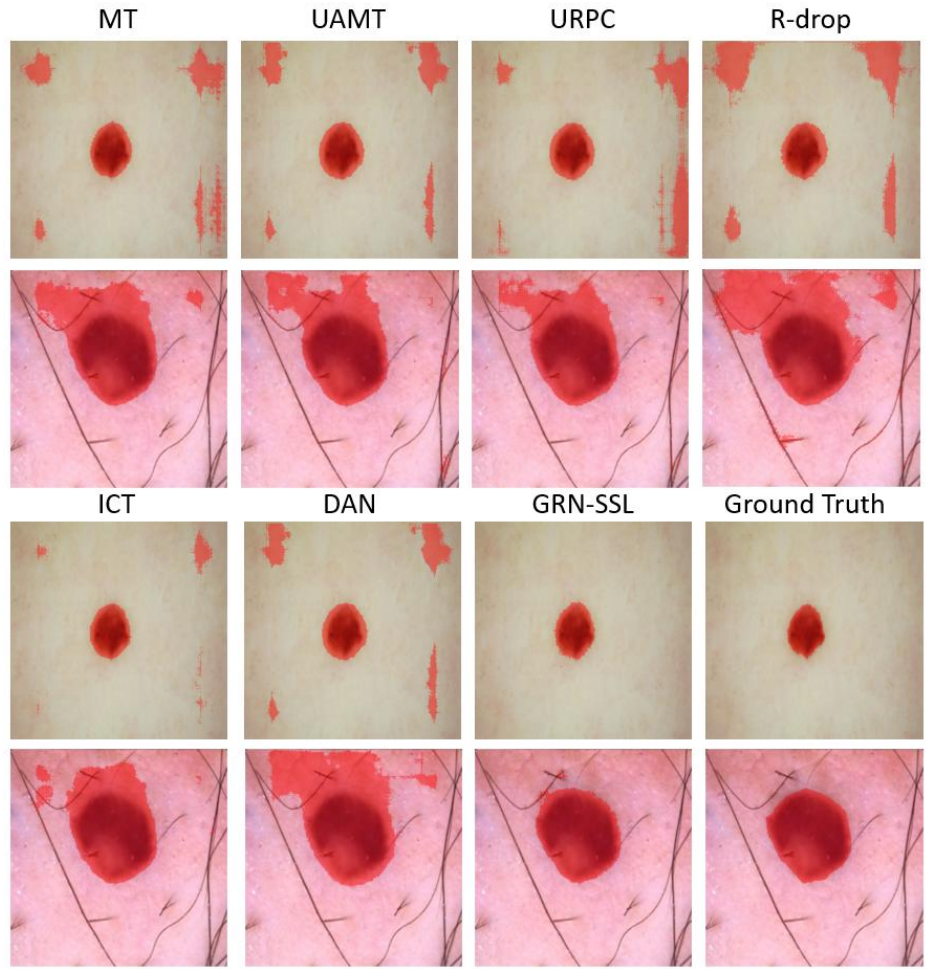

**Fig. 11**. Mask predictions for GRN methods and other methods when trained on the 5% labeled images in the demoscopic skin-lesion dataset. GRN-SSL could more accurately predict the skin-lesion regions than other SSL methods.

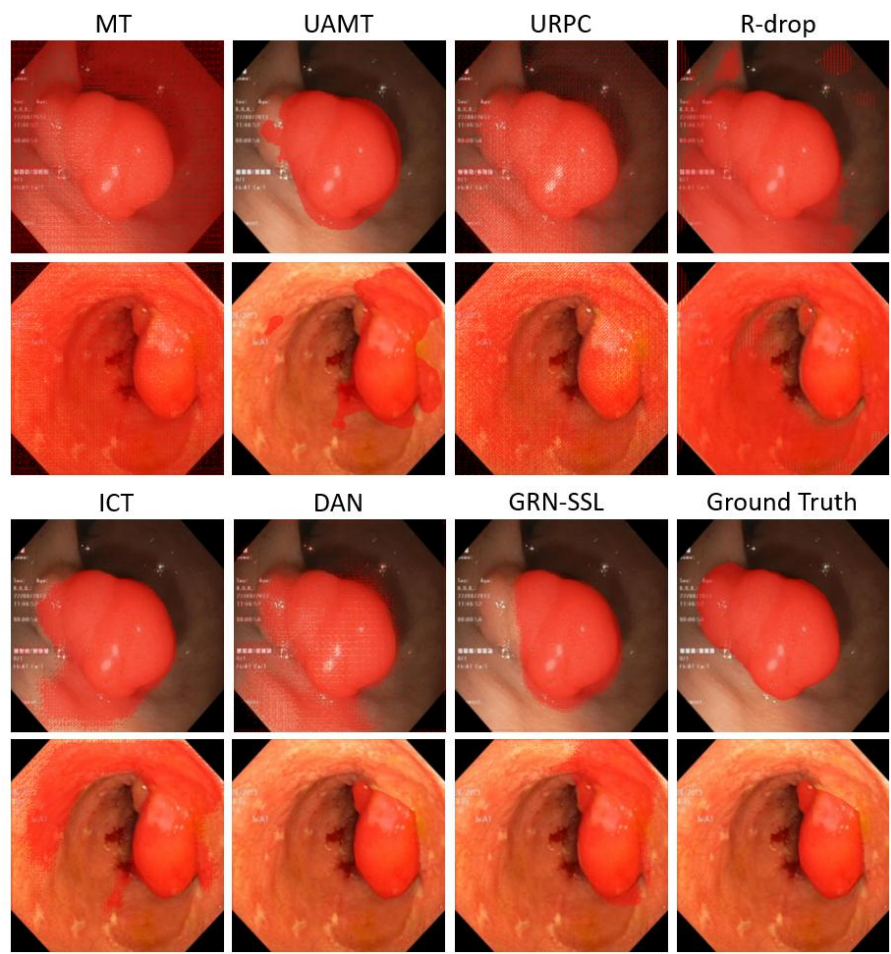

**Fig. 12**. Mask predictions for GRN methods and other methods when trained on the 5% labeled images in the Kvasir dataset.

### 3.5 Ablation Study

Table 7 presents the results of the ablation study on the GRN framework. The findings indicate that neither standard image augmentation nor traditional GAN augmentation methods achieved a DSC as high as GRN-SEL with DGA. The inclusion of SGE yielded a marginal increase of 0.18%. The removal of $L_{seg}$ feedback from GRN-SEL led to a significant 5.76% decrease in DSC. The negated $L_{seg}$ feedback further significantly degrades the segmentor's performance. Additionally, freezing the segmentor weights during the backpropagation of the generator loss $L_G$ caused a further 8.56% decline in DSC. Conversely, implementing interpolation consistency loss with unlabeled data in unsupervised training resulted in the greatest improvement, increasing DSC by 25.08%.

**Table 7**. Results of the ablation study evaluating the proposed GRN framework on the validation dataset when 5% of images being labeled.

| Method | DSC(%) | Improvement(%) |
| --- | --- | --- |
| Fully Supervised | 40.02 (39.49, 40.55) | - |
| Fully Supervised (w. image augmentation) | 38.15 (37.78, 38.52) | -1.87 |
| Fully Supervised (w. GAN augmentation) | 53.82 (53.11, 54.53) | +13.80 |
| GRN-SEL | 57.12 (56.19, 58.05) | +17.10 |
| GRN-SEL (w. SGE) | 57.30 (56.44, 58.16) | +17.28 |
| GRN-SEL (w/o $L_{seg}$ feedback) | 51.36 (50.38, 52.34) | +11.34 |
| GRN-SEL (w. -$L_{seg}$ feedback) | 32.74 (32.14, 33.34) | -7.28 |
| GRN-SEL (Freeze S for back-propagate $L_G$) | 48.56 (47.45, 49.67) | +8.54 |

Segmentation performance are presented as percentage ± half of CI. Segmentation performance are calculated excluding the background class. Image augmentation includes geometric augmentation and intensity augmentation. GAN augmentation indicates train GAN model for reconstructed purpose and implement the pre-trained generator to generate joint dataset ($D_{original}, D_{reconstructed}$) for segmentor to trained with. $L_{seg}$ feedback indicates including segmentation loss $L_{seg}$ term in generator loss $L_G$.

## 4. Discussion

We developed and validated a novel method called GRN, which synergistically integrates the joint training of a GAN model and a segmentor to accurately segment tissue layers depicted on B-model ultrasound imaging while significantly reducing the need for extensive image annotations. Unlike traditional unconstrained GAN-based data augmentation methods, our GRN framework features a feedback mechanism in which the segmentation model provides feedback loss to the generator, thereby guiding the image generation process to produce easy-to-learn segmentation-optimized images. Our results shows that GRN achieved superior performance than other existing SSL methods across various labeling levels in all three datasets. When it comes to labeling reduction, our results demonstrate that in the LBUD dataset, GRN-SSL with SGE can reduce data labeling efforts by up to 70% (Table IV). Meanwhile, GRN-SSL alone achieves a 60% reduction, GRN-SEL with SGE also reduces data labeling efforts by 70%, and GRN-SEL alone reduces by 60%, all while maintaining performance comparable to fully supervised models. Notably, GRN-SSL with SGE outperforms models trained on all labeled images, achieving a 2.59% higher DSC with only 40% of the labeled images in the training set. These findings suggest the effectiveness of the GRN framework in optimizing segmentation performance with significantly less labeled data, offering a scalable and efficient solution for medical image analysis while alleviating the burdens associated with data annotation.

The GRN framework offers several advantages. First, the generator continuously updates its weights, providing the segmentor with dynamically augmented images across different epochs. This process substantially increases the diversity of input data and finally improves the robustness of the model. Second, the generator serves as an optional image enhancement tool, specifically optimized to generate images that minimize the segmentor's loss, thereby further enhancing segmentation performance. Third, the GRN framework exhibits significant versatility. Based on the GRN-SEL, the Dice loss used to measure the performance of the segmentor could be replaced by other loss functions, such as interpolation consistency loss for GRN-SSL or rectified pyramid consistent loss in the URPC model [32, 34]. Lastly, unlike other semi-supervised learning approaches that typically require a large set of unlabeled images,

GRN-SEL only relies on a small set of annotated data and supports semi-supervised learning without special requirements for the number of unlabeled images.

Contrary to the prevailing belief that augmentation should generate challenging and diverse datasets to enhance model performance [20], our ablation study results demonstrate that using a less diverse, denoised augmented dataset can improve model performance (Table VI). This is because training on a denoised dataset allows the model to focus on informative features. Consider a sample with two feature components ($X_{informative}, X_{noise}$), and corresponding weight matrices ($W_{informative}, W_{noise}$), the extracted feature can be expressed as: $F = W_{informative} \times X_{informative} + W_{noise} \times X_{noise}$. If the noise features are set to 0, the gradient can be expressed as:

$$\frac{\partial L}{\partial W_{noise}} = \frac{\partial L}{\partial F} \times \frac{\partial F}{\partial W_{noise}} = \frac{\partial L}{\partial F} \times X_{noise} = 0$$
$$\text{when } X_{noise} = 0 \qquad (12)$$

In this scenario, the weights associated with noise features $W_{noise}$ are not updated and remain near their initial values, close to zero. This enables the model to ignore noise features and concentrate on the informative parts of the data, thereby improving overall performance.

We did not freeze segmentor weights during the backpropagation of the generator loss $L_G$. Although GRN shares a feedback mechanism similar to GANs by allowing the generator to receive guidance, the role of the discriminator and segmentor diverse in GRN differ from those in conventional GAN architectures. In GRN, the objective function is $min_G \left( min_S L_{seg}(S(G(x)), y) \right)$, where the generator $G$ and segmentor $S$ work together to minimize the segmentation loss. Additionally, during the inference stage, the generator can serve as an image enhancement module, preprocessing input images before they enter the segmentor. Therefore, we opted not to freeze the segmentor weights. This design allows both the generator $G$ and segmentor $S$ to adapt dynamically with each training iteration.

The component analysis showed that the DGA significantly improves model performance when the proportion of labeled data is low; however, its impact diminishes as the amount of labeled data in the training set increases (Table V). One explanation is when dataset diversity is limited, even a small amount of augmentation can substantially increase diversity and thereby improve model performance. As the labeled data becomes more abundant, each additional augmentation contributes less to diversity, leading to diminishing returns and a plateau in performance gains. In contrast, SGE consistently enhances model performance across all levels of labeled data, including scenarios with a fully labeled dataset. This finding suggests that beyond optimizing the encoder and decoder architectures for improved performance, incorporating a pre-enhancement model specifically tuned to minimize output loss is a vital strategy for improving model efficacy.

Our study has limitations. First, the dataset size is small due to the slow progress of patient recruitment and ongoing data collection. This limitation motivates us to develop novel solutions that minimize the need for a large dataset without spending significant time on manual annotations. To further address this concern, we set aside a relatively large independent test dataset consisting of 6430 images from 23 scans acquired on different locations of 11 subjects. Second, the DGA does not improve the model performance when more labeled images are included for training. (Table V) This may be attributed to the dataset's diversity, which is adequate for the model to learn relevant patterns, making additional augmented images less impactful on performance. Our experiments demonstrate that DGA is particularly effective when the sample size of the annotated training dataset is limited. Furthermore, the integration of SGE requires the generator to preprocess the image initially, leading to additional computational demands during the inference stage. However, SGE is an optional tool, providing users the flexibility to enhance model performance. Notably, both GRN-SEL and GRN-SSL prove their effectiveness without relying on SGE, with GRN-SSL outperforming other methods even in the absence of SGE. Despite these limitations, our experiments demonstrated the potential of the GRN in reducing the effort required for manual annotation.

## 5. Conclusion

We introduce a segmentation-aware joint training framework called GRN designed to segment tissue layers in 3-D B-mode ultrasound images. This innovative approach incorporates a segmentation loss feedback mechanism that enables the generator to produce reconstructed images specifically optimized to enhance the performance of the segmentation model. Experiments across three datasets confirm the effectiveness of GRN, showing superior performance over several existing semi-supervised learning approaches, particularly when labeled data is scarce. Additionally, our model achieves a higher performance compared to models trained on 100% labeled datasets, all while reducing the data labeling efforts by 60%. As a semi-supervised learning method, GRN relies on a small set of annotated data and can effectively utilize available unlabeled images, even when their quantity is limited.

**Declaration of generative AI and AI-assisted technologies in the writing process.**

During the preparation of this work the author(s) used ChatGPT in order to enhance manuscript readability. After using this tool/service, the author(s) reviewed and edited the content as needed and take(s) full responsibility for the content of the published article.